\documentclass[conference]{IEEEtran}
\IEEEoverridecommandlockouts
\usepackage{cite} 
\usepackage{amsmath,amssymb,amsfonts}
\usepackage{algorithm}
\usepackage{algorithmic}
\usepackage{graphicx}
\usepackage{textcomp}
\usepackage{xcolor}
\def\BibTeX{{\rm B\kern-.05em{\sc i\kern-.025em b}\kern-.08em
    T\kern-.1667em\lower.7ex\hbox{E}\kern-.125emX}}
    
\usepackage{amsthm} 
\usepackage[numbers]{natbib}
\usepackage{microtype}
\usepackage{subfigure}
\usepackage{enumitem}
\setlist{nosep}
\usepackage{booktabs} 
\usepackage{hyperref}

\usepackage{bm}
\usepackage{url}
\usepackage{float}
\usepackage{mathtools}
\usepackage[flushleft]{threeparttable} 
\newtheorem{innercustomgeneric}{\customgenericname}
\providecommand{\customgenericname}{}
\newcommand{\newcustomtheorem}[2]{%
  \newenvironment{#1}[1]
  {%
   \renewcommand\customgenericname{#2}%
   \renewcommand\theinnercustomgeneric{##1}%
   \innercustomgeneric
  }
  {\endinnercustomgeneric}
}
\newcustomtheorem{customdf}{Definition}
\newcustomtheorem{customthm}{Theorem}
\newcustomtheorem{customlemma}{Lemma}
\renewcommand{\baselinestretch}{0.98}

\usepackage[
singlelinecheck=false 
]{caption}
\usepackage{etoolbox}
\makeatletter
\patchcmd{\@makecaption}
  {\scshape}
  {}
  {}
  {}
\makeatletter
\patchcmd{\@makecaption}
  {\\}
  {.\ }
  {}
  {}
\makeatother
\def\tablename{Table}

\usepackage[subtle]{savetrees}
\renewcommand{\baselinestretch}{0.97}

\usepackage[absolute,overlay]{textpos}
\begin{document}

\begin{textblock*}{\textwidth}(3.5cm,0.3cm)
\large
To Appear in the 2020 IEEE International Conference on Data Mining (ICDM).
\end{textblock*}

\title{LFGCN: Levitating over Graphs with Levy Flights
}

\author{\IEEEauthorblockN{Yuzhou Chen}
\IEEEauthorblockA{\textit{Department of Statistical Science} \\
\textit{Southern Methodist University}\\
Dallas, USA \\
yuzhouc@smu.edu}
\and
\IEEEauthorblockN{Yulia R. Gel}
\IEEEauthorblockA{\textit{Department of Mathematical Sciences} \\
\textit{University of Texas at Dallas}\\
Richardson, USA \\
ygl@utdallas.edu}
\and
\IEEEauthorblockN{Konstantin Avrachenkov}
\IEEEauthorblockA{\textit{Network Engineering and Operations} \\
\textit{Inria Sophia Antipolis}\\
Sophia Antipolis, France \\
k.avrachenkov@inria.fr}
}

\maketitle

\begin{abstract}
Due to high utility in many applications, from social networks to blockchain to power grids, deep learning on non-Euclidean objects such as graphs and manifolds, coined Geometric Deep Learning (GDL), continues to gain an ever increasing interest. We propose a new L\'evy Flights Graph Convolutional Networks (LFGCN) method for semi-supervised learning, which casts the L\'evy Flights into random walks on graphs and, as a result, allows both to accurately account for the intrinsic graph topology and to substantially improve classification performance, especially for heterogeneous graphs. Furthermore, we propose a new preferential P-DropEdge method based on the Girvan-Newman argument. That is, in contrast to uniform removing of edges as in DropEdge, following the Girvan-Newman algorithm, we detect network periphery structures using information on edge betweenness and then remove edges according to their betweenness centrality. Our experimental results on semi-supervised node classification tasks demonstrate that the LFGCN coupled with P-DropEdge accelerates the training task, increases stability and further improves predictive accuracy of learned graph topology structure. Finally, in our case studies we bring the machinery of LFGCN and other deep networks tools to analysis of power grid networks -- the area where the utility of GDL remains untapped.
\end{abstract}

\begin{IEEEkeywords}
graph-based semi-supervised learning, convolutional networks, 
L\'evy flights, local graph topology
\end{IEEEkeywords}

\section{Introduction}
\label{Introduction}
Adaptation of deep learning (DL) to graphs and other non-Euclidean objects has recently witnessed an ever increasing interest, leading to the new subfield of {\it geometric deep learning (GDL)}. In particular, geometric deep learning is an emerging direction in machine learning which generalizes concepts of deep learning for data in non-Euclidean spaces, e.g., graphs and manifolds, by bridging the gap between graph theory and deep neural networks~\citep{bronstein2017geometric, monti2017geometric, GDL}. 

Many such DL approaches for non-Euclidian objects are based on the idea of a convolution operation in the spectral domain with a suitably chosen nonlinear trainable filter~\citep[see an overview by][]{wu2020comprehensive}. As a result, 
node features are mapped into some Euclidian space. Next, graph filters are approximated with various finite order polynomials, e.g., Chebyshev polynomials~\citep[i.e., the ChebNet model family of][]{defferrard2016convolutional, kipf2017}, Cayley transform ~\citep[i.e., CayleyNet of][]{levie2018cayleynets} or the generalization of polynomial filters in a form of Auto-Regressive Moving Average (ARMA) models~\citep{bianchi2019graph}. 
However, deep learning approaches based on approximation with finite order polynomials tend to be non-robust to even minor changes in the graph structure and to largely disregard the local graph topology which often plays the critical role for learning on heterogeneous graphs. In contrast, as noted by~\citep{bianchi2019graph}, one of the primary benefits of the ARMA filters over polynomial ones is that ARMA filters are not computed in the Fourier space induced by a graph Laplacian, and as a result, ARMA filters are local in the node space and enable to more flexibly and accurately capture the underlying graph topology. 

Further advancing the localized approaches in GDL, we propose a new fractional generalized graph-based convolutional filter for semi-supervised learning which casts the L\'evy Flights into random walks on graphs. As a result,
our new L\'evy Flights Graph Convolutional Network (LFGCN) method allows to more accurately account for the intrinsic local graph topology and to substantially improve classification performance, especially for heterogeneous graphs.

{\bf To fly or not to fly, and if to fly, why take  a L\'evy Flight?} L\'evy Flight is a random process with a scale-free, L\'evy stable jump length distribution. Due to the scale-free character, throughout the graph exploration we move randomly according to a power-law distribution for hops, rather than with integer hops as in a standard random walk. As a result, L\'evy Flight  delivers more accurate and efficient search strategies, especially, in sparse environments, comparing to other types of random walks~\citep{riascos2012long}. While superiority of
L\'evy Flights as a primary search strategy has been proven in a broad range of settings, utility of L\'evy Flights in graph-based learning remains unexplored. 
Hence, L\'evy Flights offer a new learning perspective with multiple advantages comparing to the currently available architectures.
First, due to a fractal character, L\'evy Flights combines both local graph exploration and long-range excursions, which reduces oversampling comparing to a normal random walk (i.e., lower probability to revisit the nodes we have already seen). Second, 
L\'evy Flights allow to directly reach long-distance nodes without intervention of intermediate nodes.
Third, based on average global time, L\'evy Flights average return probability is of power-law distr., i.e. $p_0^{\gamma} (t) \sim t^{-1/2\gamma}$ and is lower than average return probability of a normal random walk $p_0^{1} (t) \sim t^{-1/2}$, thereby leading to more efficient graph exploration.
Forth, L\'evy Flights are known to exhibit a particularly high utility for unbalanced and directed data which can explain higher LFGCN accuracy we have obtained in directed networks.

In addition, to abate over-fitting and over-smoothing in GDL, we develop a new preferential P-DropEdge method based on censoring edge order statistics at each training epoch. Our P-DropEdge idea is inspired by the recent 
DropEdge algorithm of~\citep{rong2019truly} 
and is rooted in nonparametric methods, specifically, various censoring schemes, for statistical inference on order statistics~\citep{balakrishnan2014art}. 
In contrast to uniformly removing edges as in the recent
DropEdge algorithm of~\citep{rong2019truly}, we follow the Girvan-Newman argument and target edges
that tend to contribute more to the intrinsic graph topology. That is, we randomly remove
edges with higher betweenness centrality, or the corresponding higher edge order statistics.
The intuition is the following.
In both P-DropEdge and DropEdge the goal is to introduce randomness in the network structure. If we are to learn international political networks with GDL, DropEdge largely tends to remove connections among individual citizens while P-DropEdge randomly censors collaboration links among Presidents and Prime Ministers. Removal of such targeted connection is likely to lead to higher perturbation effects.  
We investigate utility of the new P-DropEdge approach vs. DropEdge in conjunction with LFGCN
and GMMN (the best performing baseline) of~\citep{qu2019gmnn}.

\vskip 0.02in
Significance of our contributions can be summarized as:
\begin{itemize}[leftmargin=*]
    \vskip 0.022in
    \item we propose a new fractional generalized graph-based convolutional filter for semi-supervised learning which casts the L\'evy Flights into random walks on graphs and, as a result, provides a more efficient exploration of graph structures. We develop a new L\'evy Flights Graph Convolutional Network (LFGCN) method that substantially improves accuracy of node classification for graphs, outperforming
    12 State-of-the-Art (SOTA) methods on 2 out 3 considered undirected networks and 10 SOTAs on all 4 considered directed networks.  
    
    \vskip 0.022in
    \item the proposed architecture of LFGCN uses three state-of-the-art operations -- \textit{gated max-average pooling}, residual block, and P-DropEdge. 
    We provide an ablation study and investigate contribution of each component to the resulting classification accuracy as well as explore sensitivity of the overall system architecture to (hyper)parameter settings. 
    
    \vskip 0.022in
    \item we provide theoretical foundations behind the proposed LFGCN architecture and show that the proposed LFGCN architecture leads to significant gains in the training convergence and model output stability.
    
    \vskip 0.022in
    \item the developed preferential P-DropEdge based on censoring of higher edge betweenness order statistics is shown to exhibit utility in other GCN methods and, hence, might be applicable in broader GDL settings. 
    
    \vskip 0.022in
    \item Last but not the least, while validating our LFGCN methodology, we bring the GDL concepts to the analysis of power grid networks, i.e., the area of critical societal importance where to the best of our knowledge, the GDL machinery has never been yet applied.    
    
\end{itemize}

\section{Related Work}
\label{related_works}
Many earlier semi-supervised learning approaches on graphs, e.g., Gaussian mixture models, co-training, harmonic function, and label propagation, tend to employ only the label information (i.e., labeled instances) for training models based on the smoothness assumption over the labels~\citep{zhu2009introduction} and to largely disregard the underlying graph structure. To enhance performance, several learning methods on graphs propose to incorporate intrinsic ``graph-based'' information by designing a classifying function via generalizing the normalized cut and adding a smooth function with respect to the intrinsic structure~\citep{zhou2004learning, zhou2007spectral}. 
An optimization framework of~\citep{avrachenkov2012generalized} generalizes these approaches by considering the above two methods as particular cases. However, the major criticism to these graph-based semi-supervised learning methods is  that important information contained in graph edges is largely disregarded.

To address these limitations, \citep{defferrard2016convolutional} propose a formulation of convolutional neural networks (CNN) based on spectral graph theory -- ChebNet. ChebNet employs approximation via finite order polynomials and is based on the Chebyshev expansion for fast filtering instead of the expensive eigen-decomposition. 
Graph Convolutional Networks (GCN) of~\citep{kipf2017} simplifies ChebNet while further addressing the gradient vanishing problem and reducing the number of optimization. Other related approaches to graph learning with deep neural networks include, for instance, mixture model networks (MoNet)~\citep{monti2017geometric},
graph attention networks (GAT)~\citep{velivckovic2017graph}, graph convolutional recurrent networks~\citep{seo2018structured}, dual graph convolutional networks~\citep{zhuang2018dual}, FastGCN~\citep{chen2018fastgcn}, and simplified version of GCN~\citep{wu2019simplifying}. By directly powering the graph Laplacian, GCN based on random walks such as approximate personalized propagation of neural predictions (APPNP)~\citep{klicpera2018predict}, variable power network (VPN)~\citep{jin2019power}, and MixHop~\citep{abu2019mixhop} can learn the relationships between multiple-hops neighborhood.

To extend the success of GCN on undirected graphs to directed graphs, MotifNet of~\citep{monti2018motifnet} replaces the normalized Laplacian with the \textit{motif Laplacian} in a multivariate polynomial filter, where the motifs information can help capture the network structure. 
Finally, the most recent approach of~\citep{bianchi2019graph} provides more flexible responses than GCN by using parallel and periodic concatenations of the convolutional kernel via the ARMA filter. As a result, the ARMA approach which is applicable to both directed and undirected networks allows to more accurately incorporate the underlying local graph structure into the graph learning process.
For a recent comprehensive overview of GCNs see~\citep{wu2020comprehensive}.
\section{Methodology}
\label{methodology}
Consider a graph structure $\mathcal{G} = \{\mathcal{V}, \mathcal{E}, W$\}, where $\mathcal{V}$ is a node set with
cardinality $|\mathcal{V}|$  of $N$,
and $\mathcal{E} \subseteq \mathcal{V} \times \mathcal{V}$ is an edge set. An $N\times N$-matrix $W$
with entries $\{\omega_{ij}\}_{1\leq i,j\leq N}$ represents the adjacency matrix of $\mathcal{G}$, that is, 
$\omega_{ij} \neq 0$ for any $e_{i j} \in \mathcal{E}$ and $\omega_{ij} = 0$, otherwise. For an undirected graph $\mathcal{G}$,
$W=W^{\top}$. In reality, however, undirected graphs are often simplified representations of complex directed networks. If $\mathcal{G}$ is directed, we substitute $W$ with $W^{'} = (W^{\top}+W)/2$.

Let $Q, Q\in \mathbb{Z}_{> 0}$ be the number of different node features associated each node $v\in \mathcal{V}$. Then, a $N \times Q$ feature matrix $X$ serves as an input to an semi-supervised learning algorithm. To classify $N$ data points into $K$ classes (communities), we define a $N \times K$ label matrix $Y$ 
such that $Y_{i k}=1$ if 
vertex $i$ is labeled as class $k$, and 0 otherwise.
Here we refer to each column $Y_{\cdot k}$ of matrix $Y$ as a \textit{labeling function}. Finally, we define an $N \times K$ matrix $F$ whose columns $F_{\cdot k}$ are referred to as \textit{classification functions}.

\subsection{Graph signal processing}
Given the adjacency matrix $W$ of $\mathcal{G}$, let $D$ be the degree matrix where $d_{i i}=\sum_{j=1}^{N} w_{i j}$ and $L=U^{\top}\Lambda U$ be the Standard Laplacian matrix. Here $\Lambda = diag(\lambda_0, \dots, \lambda_{N-1})$ and $U = \left[u_{0}, \dots, u_{N-1}\right]$ is the matrix of eigenvectors.

In the following, we will revisit three popular semi-supervised learning methods - graph-based semi-supervised learning, fractional graph-based semi-supervised learning, and graph convolutional networks and gain new insights for improving their modeling capabilities.

\textbf{Graph-based semi-supervised learning} Graph-based semi-supervised learning (G-SSL) has received much attention as an alternative approach to the population paradigm of supervised learning in recent years. G-SSL develops a generalized optimization framework, which has three particular cases (i) the Standard Laplacian (SL); (ii) Normalized Laplacian (NL); (iii) PageRank (PR). The general idea of graph-based semi-supervised learning (G-SSL) is based on two widely used optimization frameworks. The first formulation, the SL based formulation~\citep{zhou2007spectral} as follows:
\begin{equation*}
    \min _{F}\left\{\sum_{i=1}^{N} \sum_{j=1}^{N} w_{i j}\left\|F_{i .}-F_{j .}\right\|^{2}+\mu \sum_{i=1}^{N} d_{i}\left\|F_{i .}-Y_{i .}\right\|^{2}\right\},
\end{equation*}
where $d_{ii}$ is $(i,i)$-element in degree matrix $D$ and $w_{ij}$ represents the edge weight for edge $e_{ij}$ in adjacency matrix $W$. For the second formulation, the NL based formulation~\citep{zhou2004learning}, is as follows:
\begin{equation*}
    \min _{F}\left\{\sum_{i=1}^{N} \sum_{j=1}^{N} w_{i j}\left\|\frac{F_{i .}}{\sqrt{d_{i i}}}-\frac{F_{j .}}{\sqrt{d_{j j}}}\right\|^{2}+\mu \sum_{i=1}^{N}\left\|F_{i .}-Y_{i .}\right\|^{2}\right\}
\end{equation*}

The following lemma~\citep{avrachenkov2012generalized} asserts that the generalized optimization framework, i.e., G-SSL, which has as particular cases the two above mentioned formulations:
\begin{customlemma}{1}
Let $\sigma$ denote an alternative parameter on the power of degree matrix $D$ whose entries are the degrees $d_{ii}$; and let $0 \leq \sigma \leq 1$. Then
\label{gssl_formula}
\begin{eqnarray*}
    \min _{F}\biggl\{\sum_{i=1}^{N} \sum_{j=1}^{N} w_{i j}
    \biggl|d_{i i}^{\sigma-1} F_{i .} - d_{j j}^{\sigma-1} F_{j .}\biggr|^{2} \\ 
    +\mu \sum_{i=1}^{N} d_{i i}^{2 \sigma-1}\left\|F_{i .}-Y_{i.}\right\|^{2}\biggr\}.
 \end{eqnarray*}
\end{customlemma}
The classification functions for the generalized semi-supervised learning are given by
$F_{\cdot k}=(1-\alpha)\left(I-\alpha D^{-\sigma} W D^{\sigma-1}\right)^{-1} Y_{\cdot k}$. 

{\it Proof} of \textbf{Lemma~\ref{gssl_formula}} is in Appendix~\ref{lemma3_proof}. The optimization formulation $S(F)$ with the following expression:
\begin{eqnarray}
\label{eq2}
    S(F) &=& \min _{F}\biggl\{2 F_{\cdot k}^{T} D^{\sigma-1} L D^{\sigma-1} F_{\cdot k}\\
    &+&\mu(F_{\cdot k}-Y_{\cdot k})^{T} D^{2 \sigma-1}(F_{\cdot k}-Y_{\cdot k})\biggr\}, \nonumber
\end{eqnarray}
where $\mu$ is a regularization parameter. Minimization of the 1st term in~(\ref{eq2}) corresponds to the idea that if two nodes are close in graph with respect to some metric, they should belong to the same class; and by minimizing the 2nd term we aim to bring the classification function $F_{\cdot k}$ as close as possible to the labeling function $Y_{\cdot k}$.  Eq.~(\ref{eq2}) allows us to obtain the Standard Laplacian based formulation ($\sigma = 1$), the Normalized Laplacian formulation ($\sigma = 0.5$), and PageRank formulation ($\sigma = 0$). Objective of the generalized optimization framework for G-SSL is a convex function and the corresponding classification function:
\begin{equation}
\label{eq_gssl}
    F_{\cdot k}=\frac{1-\alpha}{I-\alpha D^{-\sigma} W D^{\sigma-1}} Y_{\cdot k},
    \quad \alpha = \frac{2}{2+\mu}, 1\leq k \leq K.
\end{equation}

By tuning the parameter $\sigma$ on the power of degree matrix $D$, we can obtain three mentioned above particular semi-supervised learning methods:
\begin{eqnarray*}
\sigma &=&\textbf{1} \textbf{ \medskip - \medskip} \textbf{SL}: F_{\cdot k}=(1-\alpha)\left(I-\alpha D^{-1} W\right)^{-1} Y_{\cdot k},\\
\sigma &=&\mathbf{\frac{1}{2}} \textbf{ \medskip - \medskip} \textbf{NL}: F_{\cdot k}=(1-\alpha)\left(I-\alpha D^{\frac{-1}{2}} W D^{\frac{-1}{2}}\right)^{-1} Y_{\cdot k},\\
\sigma &=&\textbf{0} \textbf{ \medskip - \medskip} \textbf{PR}: F_{\cdot k}=(1-\alpha)\left(I-\alpha W D^{-1}\right)^{-1} Y_{\cdot k}.
\end{eqnarray*}
From above formulations,  classification function $F$ is a closed form solution based on the theory of random walks on graphs, which in turn provides connection to the probabilistic interpretation of G-SSL.
Parameter $\alpha$ controls the strength of the ground truth label matrix $Y$ in the generalized optimization framework. 

\textbf{Fractional graph-based semi-supervised learning} To improve classification performance (in particular, fuzzy graphs and unbalanced labeled data) of G-SSL, fractional graph-based semi-supervised learning~\citep{de2017fractional} embeds L\'evy Flights into random walks on graphs by constructing from powers of the Laplacian matrix, i.e., the $L^{\gamma}$ operator. This operation can be used to generate different transition probabilities (i.e., corresponding to stochastic adjacency matrix) based on different $\gamma$ values. Intuitively, embedding L\'evy Flights into random walks allows for better capturing mixing properties (i.e., dependence) in the data. Based on a fractional Laplacian matrix, $0<\gamma \leq 1$, the anomalous (fractional) diffusion processes on networks can be constructed from the spectra data and eigenvectors of the Laplacian matrix. The fractional powers of $L$ allows L\'evy random walks with long-range navigation on a network. For example, the long-range transitions on a network can directly move node $u$ and node $v$ with the transition probability $m^{(\gamma)}_{u \rightarrow v}$ through a random walker, where $m^{(\gamma)}_{u \rightarrow v}$ is an element in the fractional transition matrix $\mathbf{M}^{(\gamma)}$. Transition probability $m^{(\gamma)}_{u \rightarrow v}$ between any two nodes whose geodesic distance is not infinite can be summarized as follow:
\begin{equation}
\label{prob_jump}
    m_{u \rightarrow v}^{(\gamma)}=\delta_{uv}-{\left(L^{\gamma}\right)_{uv}}/{k_{u}^{(\gamma)}},
\end{equation}
where $\delta_{uv}$ is the Kronecker delta, $k_{u}^{(\gamma)}$ denotes the fractional degree of the node $u$ and $k_{u}^{(\gamma)} \equiv\left(L^{\gamma}\right)_{uu}$. Eq.~(\ref{prob_jump}) provides transition probabilities for the L\'evy Flights. Unlike the standard random walk, the L\'evy Flights can jump immediately over several hops in a graph. This feature enables L\'evy Flights to be a very effective graph exploratory process. Lemma~\ref{lemma_eq} makes this statement formal. 

\begin{customlemma}{2}\label{lemma_eq}
The L\'evy flight defined by the normalized Laplacian has a shorter relaxation time (measure of the transience) in comparison with the original random walk.
\end{customlemma}

{\it Proof} of \textbf{Lemma~\ref{lemma_eq}} is in Appendix~\ref{lemma1_proof}. There is a price to pay for this: the typically sparse transition probability matrix becomes non-sparse. We can mitigate non-sparsity by taking a reasonable number of principal singular eigenvectors or limiting the number of terms in the Taylor expansion. Through replacing the $L$ operator with $L^{\gamma} = U^{\top}\Lambda^{\gamma} U$, the new optimization formulation $S^{*}(F)$ leaves us with the following expression:
\begin{eqnarray}
\label{fractional_closed_form}
      S^{*}(F) &=& \min _{F}\left\{2 F_{\cdot k}^{T} D_{\gamma}^{\sigma-1} L^{\gamma} D_{\gamma}^{\sigma-1} F_{\cdot k}  \right.\\
      &+&\left. \mu(F_{\cdot k}-Y_{\cdot k})^{T} D_{\gamma}^{2 \sigma-1}(F_{\cdot k}-Y_{\cdot k})\right\}, \nonumber
\end{eqnarray}
where $(D_{\gamma})_{ii} = (L^{\gamma})_{ii}$.

Let $0<\gamma \leq 1$, then the closed form solution for~(\ref{fractional_closed_form}) can be obtain as follows:
    $F_{\cdot k}=(1-\alpha)\left(I-\alpha D_{\gamma}^{-\sigma} W_{\gamma} D_{\gamma}^{\sigma-1}\right)^{-1} Y_{\cdot k}$,
for $k = 1, \dots, K$. Therefore, we can conclude three particular fractional semi-supervised learning methods like G-SSL: 
\begin{align*}
\begin{split}
\sigma &=\textbf{1} \textbf{ \medskip - \medskip} \textbf{FSL}:
F_{\cdot k}=(1-\alpha)\left(I-\alpha D_{\gamma}^{-1} W_{\gamma}\right)^{-1} Y_{\cdot k},\\
\sigma &= \mathbf{\frac{1}{2}} \textbf{ \medskip - \medskip} \textbf{FNL}:F_{\cdot k}=(1-\alpha)\left(I-\alpha D_{\gamma}^{\frac{-1}{2}} W_{\gamma} D_{\gamma}^{\frac{-1}{2}}\right)^{-1} Y_{\cdot k},\\
\sigma &=\textbf{0} \textbf{ \medskip - \medskip} \textbf{FPR}: F_{\cdot k}=(1-\alpha)\left(I-\alpha W_{\gamma}D_{\gamma}^{-1}\right)^{-1} Y_{\cdot k}.
\end{split}    
\end{align*}

\subsection{Proposed L\'evy Flights Graph Convolutional Network for semi-supervised node classification}\label{LFGCN_components}

Although both G-SSL and fractional G-SSL achieve comparable and consistent (low variance) performance on some datasets, e.g., Les Miserables, Wikipedia-math, and MNIST, these approaches consider only the given adjacency matrix $W$ and the label matrix $Y$, without using the feature matrix $X$. This limitation is crucial, especially when dealing with datasets that not only exhibit a sophisticated topological graph structure but also provide node feature information, such as citation, biological, financial, and power grid networks. To address this limitation, there have been recently proposed 
many graph-based neural networks methods, e.g., graph convolutional networks (GCN), 
which use the feature matrix $X$ instead of the label matrix $Y$ and encode the graph structure by using neural network framework. Such graph-based neural networks have been shown to achieve impressive gains in semi-supervised learning performance on graphs. Next, we turn to discussing on how the idea of L\'evy Flights can be incorporated to GCN, leading to the new L\'evy Flights Graph Convolutional Network (LFGCN) for semi-supervised node classification. 

\textbf{L\'evy Flights Graph Convolutional Network (LFGCN)} The key idea behind our proposed method is Fractional Generalized Sigma-based (FGS) filter 
$$g_{FGS}(\alpha, \sigma, \gamma) =\frac{1-\alpha}{I-\alpha D_{\gamma}^{-\sigma} W_{\gamma} D_{\gamma}^{\sigma-1}}= \frac{1-\alpha}{I - \alpha\Tilde{L}}.$$ 
To avoid the inverse computations, we insert the Taylor series expansion into the FGS filter, resulting in:
\begin{eqnarray}
\label{Taylor}
        g_{FGS}(\alpha, \sigma, \gamma)
        = (1-\alpha) \sum_{i=0}^{\infty} \bigl(\alpha \Tilde{L}\bigr)^{i},
        \quad 0< \alpha, \sigma, \gamma \leq 1.
        \end{eqnarray}
Empirically, it shows that $i = \lceil 4\alpha \rceil$ is enough to get a good approximation. We then obtain the general classification function by multiplying~(\ref{Taylor}) by the feature matrix $X$:
\begin{eqnarray}
\label{filter_expansion}
    \bar{\mathcal{X}} & =& g_{FGS}(\alpha, \sigma, \gamma) X \\
       & =& (1-\alpha)\Bigl(\underbrace{X}_\text{the 1st item} + \underbrace{ \alpha\Tilde{L}X}_\text{the 2nd item} + \underbrace{ \alpha^2\Tilde{L}^2 X}_\text{the 3rd item} + \cdots\Bigr),\nonumber \\
       & =& (1-\alpha) (X^{'})_i \nonumber
\end{eqnarray}
where 
$(X^{'})_i = X + \alpha\Tilde{L} (X^{'})_{i-1}$, $(X^{'})_{0} = X$, $i\in \mathbb{Z}_{i\geq 0}$.

\textbf{Convolutional layer} During LFGCN training, the convolutional model needs to train parameters $(\mathbf{W}, \mathbf{b})$ of the graph filter, where the trainable graph filter scan the given input feature matrix into a series of feature maps with neurons. Thereby, we provide an implementation of~(\ref{filter_expansion}) as a FGS convolutional layer:
\begin{equation}
\label{FGS_convolutional_layer}
    H^{(t+1)}=\sigma\left((1-\alpha) \sum_{i=0}^{\infty} \Big(\alpha \Tilde{L}\Big)^{i}H^{(t)} \mathbf{W}^{(t)}\right),
\end{equation}
where $H^{(t+1)}$ is the hidden layer output matrix of activations in the $t$-th layer and $H^{(0)} = X$, $\sigma (\cdot)$ is the adopted activation function, and $\mathbf{W}^{t}$ is the trainable weight in the $t$-th layer. Furthermore, we bring the concept of the parallel system (PS) from the reliability theory to improve the consistency of our proposed method. A parallel system is a configuration such that the entire system functions as long as not all involved components in the system fail. Hence, the parallel system structure is more robust against noisy inputs, compared to a single system structure.

\begin{customlemma}{3}\label{lemma1_eq}
{\it
Let $\bar{X}_{FGS}$ be the output matrix $P_1$ from a pooling layer. Let $\mathcal{U}=\{1,2,\dots, N\}$ be a finite population such that each unit $i, i\in \mathcal{U}$ is associated with an output matrix $X_{FGS}^{(i)}$, $i=1,\ldots, N$.
Then
    $\operatorname{Var}(\bar{X}_{FGS}) =\left(1-{n}/{N}\right) {S^{2}}/{n}$, $n<N, n\in \mathbb{Z}_{>0}$,
where $S^2 = \sum_{i=1}^{N}\bigl(X_{FGS}^{(i)}-\bar{X}_{FGS}^{U}\bigr)^{2} /(N-1)$.
}

Suppose there are $n$ components in a parallel system, with the probability of non-failure $P_R^{(i)}$ (where $i = 1, \cdots, n$) in a parallel system, then the reliability of this parallel system $P_R^{PS}$ can be obtained with the following expression:
\begin{equation*}
\label{stability}
    P_R^{PS} = 1 - \prod_{i=1}^{n}(1 - P_R^{(i)}).
\end{equation*}
\end{customlemma}

{\it Proof} of \textbf{Lemma~\ref{lemma1_eq}} is in Appendix~\ref{lemma2_proof}. According to Lemma~\ref{lemma1_eq}, the introduced concept of a parallel system allows for enhancing stability and reducing estimation variance up to order of $n$ (i.e., $\operatorname{Var}(\bar{X}_{FGS})= O (S^2/n)$). In this way, we establish both theoretical and practical guarantees for our proposed model to reach stable over a large set of hyperparameters, small datasets, and noisy labels based on this parallel implementation. 
\begin{figure*}[htp]
	\centering
	\includegraphics[width=0.77\textwidth]{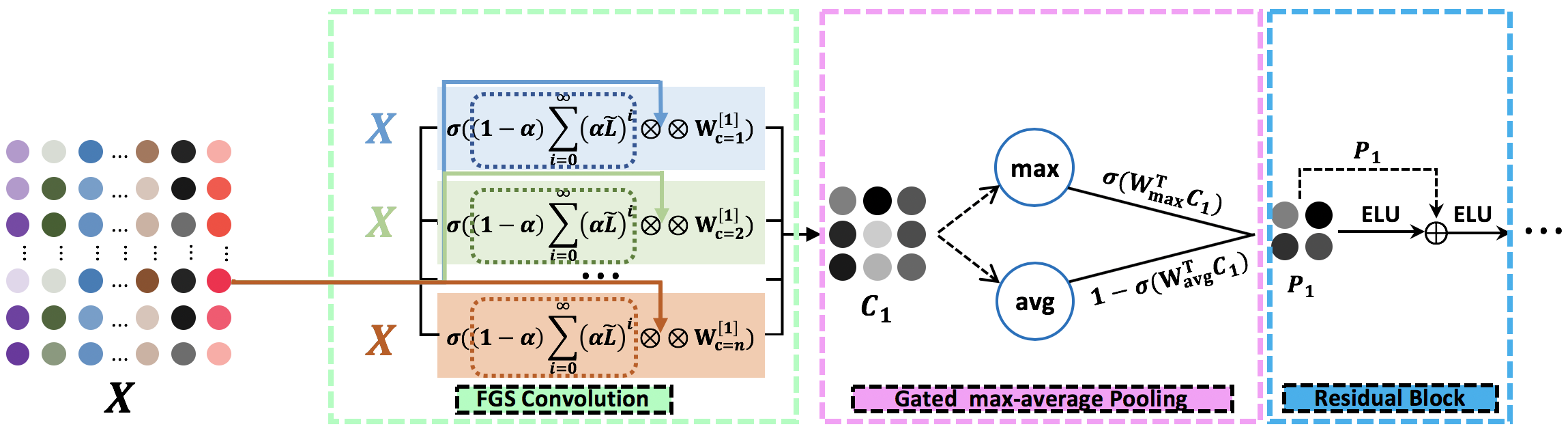}
	\caption{Illustration L\'evy Flights Graph Convolutional Network model. The input is the feature matrix $X$ and the graph within dotted circle represents embedding L\'evy Flights into random walks on graph (where $L^{\gamma}$ is the Laplacian matrix $L$ to a power $\gamma$). LFGCN architecture consists of three main components: (i) FGS convolutional layer with parallel structure; (ii) \textit{gated max-average pooling} layer; (iii) activation block for residual learning.} \label{flowchart}
	\vspace*{-0.5cm}
\end{figure*}

\textbf{Pooling layer} When implementing the form of pooling operation to aggregate information from the outputs of parallel FGS convolutional layer, instead of using some popular pooling functions such as max and average pooling, we apply the state-of-the-art pooling operation - \textit{gated max-average pooling}~\citep{lee2016generalizing} to capture the local and global information from all the nodes and graph structure. The rationale behind the \textit{gated max-average pooling}, is that it considers ``responsive'' strategy (i.e.,  improving translation invariance and scale invariance via considering input in each gating mask) based on the \textit{mixed max-average pooling} equation. That is,
\begin{eqnarray*}
    f_{\mathrm{gated}}(X_{FGS})&=&\sigma\left(\textbf{W}^{\top} X_{FGS}\right) f_{\max }(X_{FGS})\\
    &+&\left(1-\sigma\left(\textbf{W}^{\top} X_{FGS}\right)\right) f_{\mathrm{avg}}(X_{FGS}), 
\end{eqnarray*}
where $\textbf{W}$ is the trainable weight matrix, $X_{FGS}$ is the output matrix from the parallel FGS convolutional layer after concatenation operation.

\textbf{Residual building block} Inspired by the seminal works of~\citep{hamilton2017inductive, liu2019geniepath} that implemented residual learning in a graph convolutional network, we apply a residual block (RB) by adding the skip connection after the pooling layer. One of the advantages of the residual learning is the \textit{identity} mapping which provides a “direct” path for propagating information. When using the residual building block, we adopt a similar scheme as \citep{he2016deep} to deal with the output of the pooling layer. Let $\mathcal{H}(x)$ be an underlying mapping and we cast it as  $\mathcal{H}(x) = \mathcal{F}(x) + x$, where $\mathcal{F}(x)$ is the residual mapping, defined by $\mathcal{H}(x) - x$. That is, optimizing the residual mapping $\mathcal{F}(x)$ is easier than optimizing the direct mapping $\mathcal{H}(x)$ and helps to avoid the gradient vanishing problem during training. We use an exponential linear unit (ELU) in direct mapping and place a rectified linear unit (ReLU) after addition in our model. 

\textbf{P-DropEdge}
Motivated by the recent idea of message passing inference~\citep[i.e., DropEdge of][]{rong2019truly}, we develop a new preferential DropEdge approach called the
{\it P-DropEdge} which is based on censoring higher edge betweenness order statistics. 
In particular, most recently \citep{rong2019truly} propose a flexible approach, the DropEdge algorithm which
by uniformly randomly removing a
certain proportion of edges from the input graph at each training epoch, allows to
better prevent against over-fitting and to reduce the effect of over-smoothing.
The rationale behind DropEdge on introducing more randomness and deformation into the data is
intrinsically linked and complementary to the Dropout ideas of~\citep{hinton2012improving}.   
Our approach further advances DropEdge by targeting and randomly removing edges
proportionally to their betweenness centrality, i.e., preferential edge dropout of higher edge betweenness order statistics. That is, first, our idea is based on the Girvan-Newman argument of focusing on edges which tend to play a higher role in the underlying network topology~\citep{girvan2002community}. 
Second, dropout of higher edge betweenness order statistics may be viewed as a variant of recently proposed non-uniform censoring schemes for generalized order statistics in reliability theory which are shown to deliver more robust parameter estimates in heterogeneous probability distributions~\citep{balakrishnan2008progressive}. 

\begin{customdf}{1.}
(Edge Order Statistics) Given a input graph $\mathcal{G} = \{\mathcal{V}, \mathcal{E}, W$\}, the betweenness centrality for the edge $e \in \mathcal{E}$ is defined as
    $C_{B_{e}}(e)=\sum_{u \neq v \in \mathcal{V}}{\sigma_{u v}(e)}/{\sigma_{u v}}$,
where $\sigma_{uv}$ the number of shortest paths connecting $u$ to $v$, and $\sigma_{uv}(e)$ the number of shortest paths connecting $u$ to $v$ passing through the edge $e$. We then arrange edges in ascending order of their 
betweenness $\{C_{B_{e}}(e)_{i}\}$, $i = 1,2,\dots,|\mathcal{E}|$ as
    $C_{B_{e}}(e)_{(1)} \leq C_{B_{e}}(e)_{(2)} \leq \cdots \leq C_{B_{e}}(e)_{(|\mathcal{E}|)}$.
Here $C_{B_{e}}(e)_{(i)}$ is said to be the $i$th-order edge betweenness score, or the $i$th-edge betweenness order statistic.
\end{customdf}
Note that the Girvan-Newman algorithm
on edge betweenness infers the edges connecting communities, that is, the edges exhibiting a more profound role in the network organization. 
As a result, P-DropEdge offers multi-fold benefits: 
(i) it constrains direction of a random walk and
acts as a ``self-avoiding" random walk, e.g., reduces the chance of moving back to the already visited graph structure; (ii) increases variability among randomly  deformed copies of the original graph.
That is, let us consider, e.g., an international political network. Randomly removing connections among Mr. and Mrs. Smith or even US Senators from Texas and California will tend to deliver a more similar resulting graph structure than randomly removing collaboration links between Trump, Macron, Putin and Johnson. 


\begin{algorithm}[h]
\footnotesize
\caption{P-DropEdge Algorithm}
\label{P-DropEdge_alg}
\begin{algorithmic}
   \STATE {\bfseries Input:} \textbf{Data} adjacency matrix $W$, \textbf{Parameter} $p_{\text{P-D.E.}}$, $\tau$
   \FOR{$i=1$ {\bfseries to} $|\mathcal{E}|$}
    \STATE calculate  $C_{B_{e}}(e)_i$ for the edge $e_i$
   \ENDFOR
   \STATE $\triangleright$ Find order statistics of $\{C_{B_{e}}(e)_{i}\}$, $i = 1,2,\dots,\mathcal{E}$
   \STATE{} \quad \quad $C_{B_{e}}(e)_{(1)} \leq C_{B_{e}}(e)_{(2)} \leq \cdots \leq C_{B_{e}}(e)_{(|\mathcal{E}|)} $
   \STATE $\triangleright$ Draw edges 
   which correspond to the top order statistics  
   \STATE{} \quad \quad $C_{B_{e}}(e)_{((1-\tau)|\mathcal{E}|)} \leq  \cdots \leq C_{B_{e}}(e)_{(|\mathcal{E}|)} $
   \STATE $\triangleright$ Assign weights to the selected edges in 1st-round as
   \STATE $\psi_{(j)} = \frac{C_{B_{e}}(e)_{(j)}}{\sum_{j=(1-\tau)|\mathcal{E}|}^{|\mathcal{E}|} C_{B_{e}}(e)_{(j)}}$, $j = (1-\tau)|\mathcal{E}|,\dots, |\mathcal{E}|$,
   \STATE \quad \quad \quad \quad \quad $\bm{\psi} = \{\psi_{((1-\tau) |\mathcal{E}|)}, \cdots, \psi_{(|\mathcal{E}|)}\}$
   \STATE $\triangleright$ Weighted sampling without replacement
   \STATE \enskip \enskip \textit{randsample}$((1-\tau)|\mathcal{E}|:|\mathcal{E}|, p_{\text{P-D.E.}} \times \tau \times |\mathcal{E}|, \bm{\psi})$,
   \STATE \quad \quad obtain sample $\textbf{\textit{s}}$ with size $p_{\text{P-D.E.}} \times \tau \times |\mathcal{E}|$
   \STATE {\bfseries Output} new adjacency matrix $W_{p_{\text{P-D.E.}}} = W - W_{C_{B_{e}}(e) \in \textbf{\textit{s}}}$
\end{algorithmic}
\end{algorithm}

P-DropEdge method is presented in Algorithm~\ref{P-DropEdge_alg}. Given a resulting adjacency matrix $W_{p_{\text{P-D.E.}}}$ upon P-DropEdge application, a new fractional Laplacian takes the form $ \Tilde{L}_{p_{\text{P-D.E.}}} = D_{\gamma_{p_{\text{P-D.E.}}}}^{-\sigma} W_{\gamma_{p_{\text{P-D.E.}}}} D_{\gamma_{p_{\text{P-D.E.}}}}^{\sigma-1}$. Finally, we replace the fractional Laplacian $\Tilde{L}$ in~(\ref{FGS_convolutional_layer}) with $\Tilde{L}_{p_{\text{P-D.E.}}}$ for propagation and training. In validation and testing steps, P-DropEdge is not utilized.

\textbf{Advantages of LFGCN vs. Higher-order methods} Recently there has been a spike of interests to higher-order methods, that is, algorithms based on the graph convolutional layer with higher-order information in graphs, such as
APPNP~\citep{klicpera2018predict}, VPN~\citep{jin2019power}, and MixHop~\citep{abu2019mixhop}. In contrast to such higher-order graph architectures,
LFGCN offers multi-fold benefits: (i) Due to a fractal character, LF integrates local graph exploration with long-range excursions, which reduces oversampling comparing to standard random walks and allows for more efficient graph exploration; (ii) since high-order schemes~\citep{jin2019power,abu2019mixhop} are based on integer powers of Laplacian, when exploring 2-hops, 3-hops,..., $k$-hops, standard random walks employed in these higher-order methods can only describe larger scale graph structures, often resulting in very dense adjacency matrices and higher computation costs; (iii) L\'evy Flights average return probability is lower than average return probability of a normal random walk, implying more efficient graph exploration; (iv) L\'evy Flights reinforces separability of clusters, enhances performance for unbalanced data, and is known to yield better search results in directional data.
Clearly, these advantages are essential for learning graphs with higher heterogeneity, and for more homogeneous and balanced graphs, methods based on standard random walks may be a competitive alternative.

\section{Experimental Settings}

{\bf Directed and Undirected Datasets} Joining the previous works’ practice, we use three undirected citation networks benchmark datasets for semi-supervised learning evaluation, including Cora-ML (this Cora dataset consists of Machine Learning papers), CiteSeer and PubMed. We also evaluate our method on four directed networks -- Cora, IEEE 118-bus system (IEEE bus), Texas 2000-bus system (TX bus), and South Carolina 500-bus system (SC bus). The dataset statistics are summarized in Table~\ref{data_stat} (in Appendix~\ref{experiments_details}). We provide the more details about datasets description on Github in the Appendix~\ref{experiments_details}. 

\textbf{Baseline Methods} On undirected networks, we compare LFGCN with the following state-of-the-art semi-supervised classification approaches which include (i) using the label matrix as input: label propagation (LP)~\citep{zhu2002learning}; (ii) using the feature matrix as input: DeepWalk (DW)~\citep{perozzi2014deepwalk}, graph attention networks (GAT), GNN with ChebNet polynomials filter (ChebNet), GCN, GNNs with convolutional ARMA filters (ARMA), Graph Markov Neural Networks (GMNN)~\citep{qu2019gmnn}, Large-Scale Learnable Graph Convolutional Networks (LGCNs)~\citep{gao2018large}, Shortest Path Graph Attention Network (SPAGAN)~\citep{yang2019spagan}, APPNP, VPN, and MixHop. In addition, on undirected graphs, we use MotifNet, ChebNet, GCN, ARMA, GMNN, LGCNs, SPAGAN, APPNP, VPN, and MixHop as the benchmarks.

\textbf{Training Settings} Training task is done by using Adam optimizer with learning rate $lr_1 = 0.01$ for undirected networks and $lr_2 = \{0.1; 0.001\}$ for directed networks. To prevent our approach from over-fitting, we consider both adding dropout layer before two graph convolutional layers and kernel regularizers ($\ell_2$) in each layer. For undirected and directed networks: we follow the same experimental setup used in the baselines experiments to set the parameters of baselines. Parameters $p_{\text{P-D.E.}}$ and $\tau$ largely depend on the distributional properties of a network and can be estimated, e.g., via cross-validation. As a rule of thumb, we recommend $p_{\text{P-D.E.}}$ and $\tau$ of 5\% and 6\%, respectively, in larger networks of more than 2,000 nodes,  and $p_{\text{P-D.E.}}$ and $\tau$ of 1\% and 2\%, respectively, in smaller network of less than 1,000 nodes.  
The best hyperparameter configurations of LFGCN for each dataset by using standard grid search mechanism are available at Github link in Appendix~\ref{experiments_details}. 

\section{Results}
\textbf{1.\enskip Performance analysis} Tables~\ref{table_undirected_network} and~\ref{table_directed_network} report the average accuracy delivered by LFGCN and competing methods for undirected and directed networks, respectively. The best performance for each dataset is marked in bold. We find that LFGCN outperforms all competing approaches in all datasets, except for PubMed (LFGCN delivers the second best accuracy result). The improvement gain of LFGCN over the next most accurate method ranges from 0.29\% (for CiteSeer over GMNN) to 4.27\% (for directed IEEE 118-Bus over GMNN). Remarkably, methods that are applicable both to undirected and directed networks (i.e., \citep{defferrard2016convolutional,kipf2017,bianchi2019graph,qu2019gmnn,gao2018large,yang2019spagan,klicpera2018predict,jin2019power,abu2019mixhop}) tend to deliver noticeably lower accuracy results for a directed networks (especially on weighted-directed networks), while the new LFGCN method yields a more stable performance across both directed and undirected networks. 
In turn, PubMed (unweighted-undirected), GMNN outperforms LFGCN up to 2.63\%. Based on the obtain results, the new LFGCN approach tends to be the most competitive and, hence, preferred node classification method for sparser networks with higher label rates. Furthermore, the IEEE 118-Bus dataset is the smallest among the considered data, and we might expect to observe lower accuracy results for this dataset due to a limited training set. However, the accuracy yielded by LFGCN is among the highest ones across all datasets. For PubMed, it has the lightest tails for the degree distribution and a weak structural info (i.e., with very few links per node on average), thus LFGCN is not the best exploration choice.

We provide the training time per epoch on all datasets in the Appendix~\ref{experiments_details} (see Tables~\ref{sample-table}~\ref{run_time2} and~\ref{run_time3}).

\textbf{2.\enskip Ablation study by removing individual components in LFGCN} To discover the vital components in the success of our LFGCN, we investigate the contributions of individual components proposed in Section~\ref{LFGCN_components} to the performance of LFGCN. We conduct experiments by removing individual component separately (in the spirit of leave-one-out operation) from our LFGCN architecture, leading to a network without P-DropEdge, parallel structure, residual block, or \textit{gated max-average pooling}. Table~\ref{four_components_comparison} provides the comparison results between LFGCN without P-DropEdge, parallel structure, residual block, or \textit{gated max-average pooling}. The results show that LFGCN consistently outperforms the reduced LFGCN baselines by a significant
margin, reaching around 0.24\% to 2.91\% relative improvement on Cora-ML and IEEE 118-bus system. These results demonstrate contributions of all components to performance improvement.

\textbf{3.\enskip P-DropEdge vs. DropEdge} We now evaluate our P-DropEdge and regular DropEdge of~\citep{rong2019truly} based on the LFGCN and GMNN (i.e., the best performing baseline). Table~\ref{comparison_network} presents comparison between LFGCN and GMNN with regular DropEdge and P-DropEdge on CiteSeer and South Carolina 500-bus system. We find that  while a sufficiently sampling-based edge-removing is helpful for performance enhancement, regular randomly removing edges do not always improve performance. Note that this finding is in contrast to the regular DropEdge where both LFGCN and baseline equipped with P-DropEdge achieve consistently better
performance than others. These findings prove the effectiveness of employing preferential approach of P-DropEdge before the learning task.

\begin{table}[t]
\footnotesize
	\centering
		\caption{Comparison of average accuracy (\%) and standard deviation (\%) in () of semi-supervised classification approaches for undirected networks.\label{table_undirected_network}}
		\renewcommand{\arraystretch}{1.1}
	\begin{tabular}[t]{lccc}
		\toprule
		\textbf{Method}&  \textbf{Cora-ML}&\textbf{CiteSeer}&\textbf{PubMed}\\   \hline 
		LP~\citep{zhu2003semi}&68.70&46.32&65.92\\
		DW~\citep{perozzi2014deepwalk}&67.20&43.27&65.33\\
		ChebNet~\citep{defferrard2016convolutional}&81.45&70.23&78.40 \\
		GCN~\citep{kipf2017}&81.50&71.11 &79.00\\
		ARMA~\citep{bianchi2019graph}&82.80 (0.63)&72.30 (0.44)&78.80 (0.30)\\
		GAT~\citep{velivckovic2017graph}&83.11 (0.70)&70.85 (0.70)&78.56 (0.31)\\
		GMNN~\citep{qu2019gmnn}&83.72 (0.90)&73.10 (0.79)&\textbf{81.80 (0.53)}\\
		LGCNs~\citep{gao2018large}&83.35 (0.51)&73.08 (0.63)&79.51 (0.22)\\ 
		SPAGAN~\citep{yang2019spagan}&83.63 (0.55)&73.02 (0.41)&79.60 (0.40)\\ 
		APPNP~\citep{klicpera2018predict} &83.31 (0.53)&72.30 (0.51)&80.12 (0.20)\\
		VPN~\citep{jin2019power} &81.89 (0.57)&71.40 (0.32)&79.60 (0.39)\\
		MixHop~\citep{abu2019mixhop}&81.90 (0.81)&71.41 (0.40)&80.81 (0.58)\\\hline
		LFGCN w/o $p_{\text{P-D.E.}}$&84.35 (0.57)&71.89 (0.77)&79.60 (0.55)\\
		LFGCN&\textbf{84.63 (0.55)}&\textbf{73.31 (0.76)} &79.65 (0.50)\\ \bottomrule
	\end{tabular}
\end{table}%

\begin{table}[t]
\scriptsize
	\centering
		\caption{Comparison of average accuracy (\%) and standard deviation (\%) in () of semi-supervised classification approaches for directed networks.\label{table_directed_network}}
		\renewcommand{\arraystretch}{1.1}
	\begin{tabular}[t]{lcccc}
		\toprule
		\textbf{Method}&  \textbf{Cora}& \textbf{IEEE Bus}& \textbf{TX Bus}& \textbf{SC Bus}\\   \hline
		MotifNet~\citep{monti2018motifnet}&60.00&65.75&82.00&95.18\\
		ChebNet~\citep{defferrard2016convolutional}&58.93&60.00&80.04&94.13 \\
		GCN~\citep{kipf2017}&57.75&52.86&73.36&90.33\\
		ARMA~\citep{bianchi2019graph}&58.99 (0.52)&70.55 (2.23)&81.20 (0.23)&94.33 (0.47)\\
		GMNN~\citep{qu2019gmnn}&61.20 (0.50)&78.88 (2.50)&86.21 (0.29)&96.57 (0.52)\\
		LGCNs~\citep{gao2018large}&60.72 (0.43)&71.43 (2.20)&85.57 (0.25)&95.14 (0.45)\\ 
		SPAGAN~\citep{yang2019spagan}&61.00 (0.45)&78.55 (2.25)&86.00 (0.26)&96.19 (0.40)\\
		APPNP~\citep{klicpera2018predict} &61.00 (0.44)&82.05 (2.24)&86.77 (0.30)&96.11 (0.45)\\
		VPN~\citep{jin2019power} &60.53 (0.43)&82.19 (2.20)&86.87 (0.33)&96.23 (0.50)\\
		MixHop~\citep{abu2019mixhop}&60.33 (0.55)&80.05 (2.50)&86.10 (0.25)&95.94 (0.40)\\\hline
		LFGCN w/o $p_{\text{P-D.E.}}$&60.70 (0.47)&82.20 (2.30)&87.74 (0.30)&97.61 (0.47)\\
		LFGCN &\textbf{61.35 (0.45)}&\textbf{82.40 (2.24)} &\textbf{88.23 (0.26)}&\textbf{97.85 (0.47)}\\ \bottomrule
	\end{tabular}
\end{table}%

\begin{table}[t]
\footnotesize
	\centering
		\caption{Comparison of the LFGCN with/without (1) P-DropEdge ($p_{\text{P-D.E.}}$), (2) Parallel Structure (PS), (3) Residual Block (RB), and (4) \textit{gated max-average pooling} (Gated Max-Avg) in terms of node classification accuracy (\%) on undirected Cora-ML and directed IEEE118 bus system. Numbers in () are relative gains in (\%) between the best result delivered by the LFGCN and reduced LFGCN.\label{four_components_comparison}}
		\renewcommand{\arraystretch}{1.1}
	\begin{tabular}[t]{lccc}
		\toprule
		& \multicolumn{1}{c}{\textbf{Undirected}}& &\multicolumn{1}{c}{\textbf{Directed}} \\ 
		\cline{2-2}\cline{4-4}
		\textbf{Method}& \textbf{Cora-ML}&& \textbf{IEEE Bus}\\   \hline
		LFGCN w/o $p_{\text{G-N.}}$&84.35 (0.33)&&82.20 (0.24)\\
		\hline
	    LFGCN w/o PS&84.31 (0.38)&&80.71 (2.09)\\
		\hline
		LFGCN w/o RB&83.31 (1.58)&&80.07 (2.91)\\
		\hline
		LFGCN w/o Gated Max-Avg&83.66 (1.16)&&81.12 (1.58)\\
		\hline
		LFGCN &\textbf{84.63}&&\textbf{82.40}\\
		\bottomrule
	\end{tabular}
\end{table}%

\begin{table}[t]
\footnotesize
	\centering
	\caption{Comparison of GMNN and LFGCN with regular DropEdge ($p$) and P-DropEdge ($p_{\text{P-D.E.}}$) in terms of node classification accuracy (\%) on undirected Citeseer and directed South Carolina bus system. The numbers in () denote the optimal edge removal rate for the models with DropEdge and P-DropEdge.
\label{comparison_network}}
		\renewcommand{\arraystretch}{1.1}
	\begin{tabular}[t]{lccc}
		\toprule
		& \multicolumn{1}{c}{\textbf{Undirected}}& &\multicolumn{1}{c}{\textbf{Directed}} \\ 
		\cline{2-2}\cline{4-4}
		\textbf{Method}& \textbf{CiteSeer}&& \textbf{SC Bus}\\   \hline
		GMNN with $p$&73.10 ($5\%$)&&96.57 ($1\%$)\\
		GMNN with $p_{\text{P-D.E.}}$&\underline{73.20} ($5\%$)&&\underline{96.89} ($1\%$)\\
		\hline
		LFGCN with $p$&72.05 ($5\%$)&&97.58 ($1\%$)\\
		LFGCN with $p_{\text{P-D.E.}}$&\underline{\textbf{73.31}} ($5\%$)&&\underline{\textbf{97.85}} ($1\%$)\\
		\bottomrule
	\end{tabular}
\vspace*{-0.6cm}	
\end{table}%

\begin{figure}[h]
\begin{center}
\centerline{\includegraphics[width=\columnwidth]{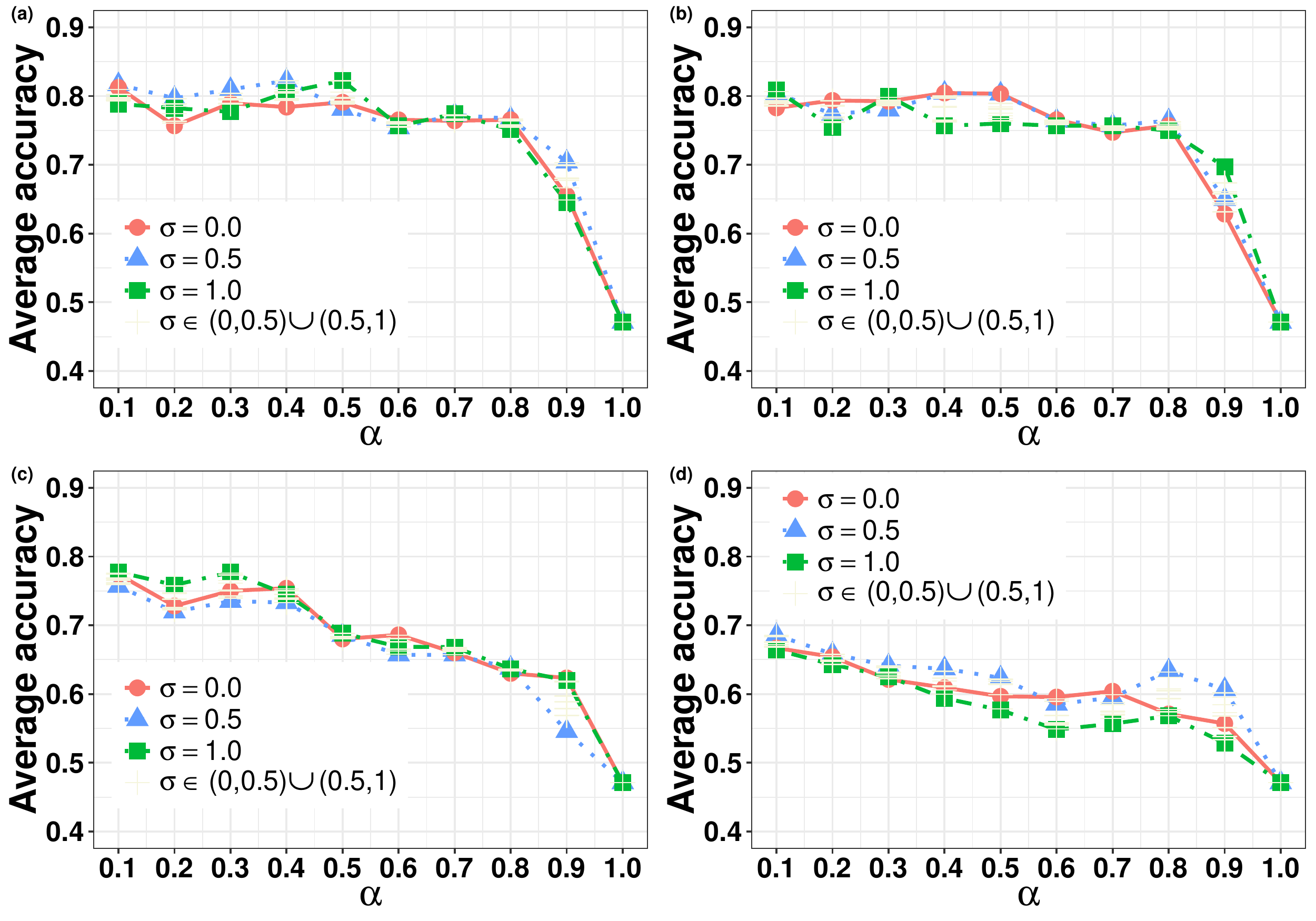}}
\vspace*{-4mm} 
\caption{Accuracy of LFGCN for PR, NL, and SL as a function of $\alpha$ and fractional power $\gamma$ = \{0.001 (a), 0.01 (b), 0.1 (c), 1.0 (d)\}.\label{different_gamma_plots}}
\end{center}
\end{figure}

\textbf{4.\enskip Evaluation of LFGCN-specific parameters} During grid search over three parameters (i.e., $\alpha$, $\sigma$, and $\gamma$), we find that:  (i) the regularization parameter $\alpha$ which used to specify the relative importance of a graph in clustering strongly relates to the probability of initial conditions for random walks when the self-refreshing process works, and it strongly influences the network's generalization ability and node classification performance for all datasets; (ii) the free unifying parameter $\sigma$ provides enough flexibility to construct a canonical formulation of different graph-based semi-supervised methods -- Table~\ref{table_undirected_network} and Table~\ref{table_directed_network} indicate that the optimal value of $\sigma$ depends on both the types of networks (undirected and directed) and label rate not on the size of network; (iii) the fractional power parameter $\gamma$ substantially impacts the accuracy of node classification for the small datasets (see e.g., Figure~\ref{different_gamma_plots}), however, no similarly strong influence is found in the larger datasets.

\begin{figure}[h]
\vskip -0.3in
\begin{center}
\centerline{\includegraphics[width=\columnwidth]{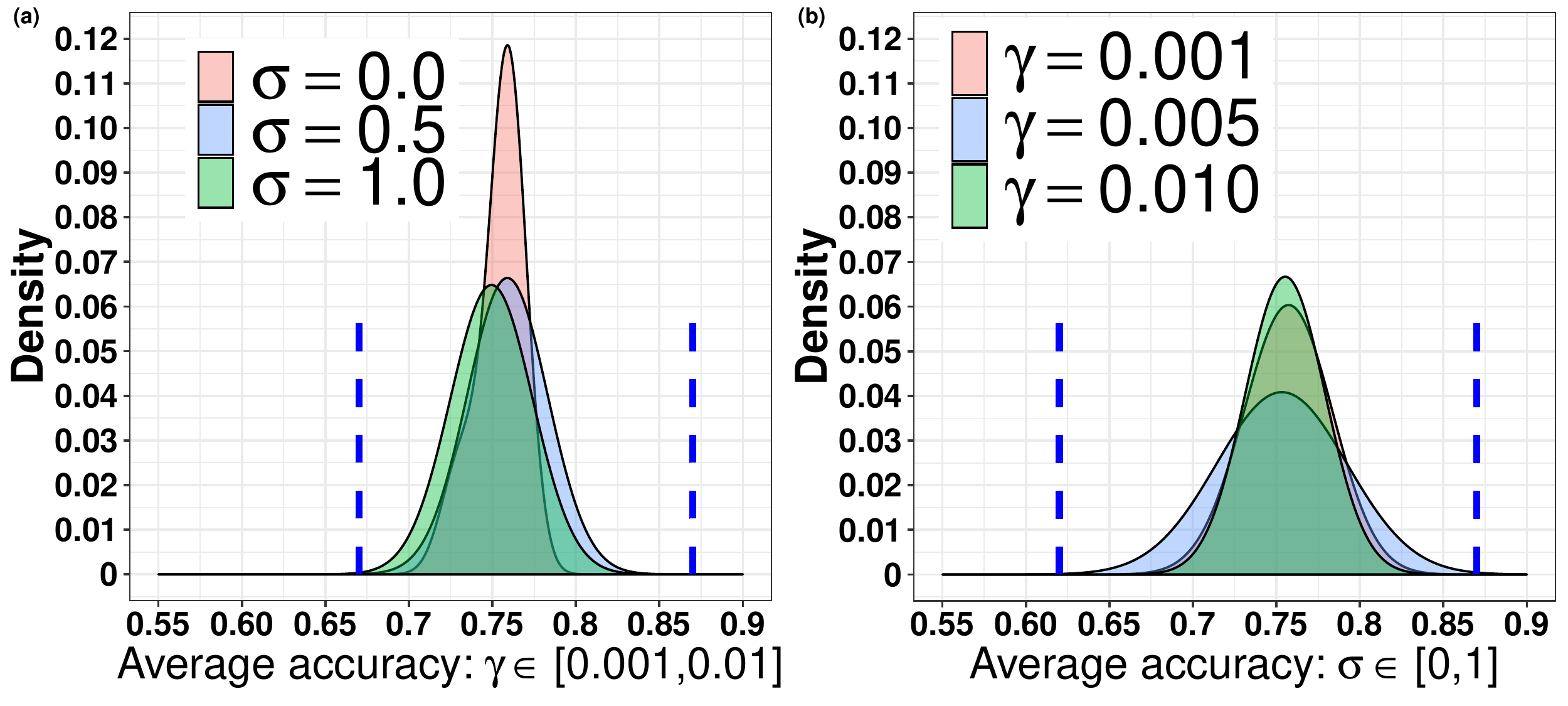}}
\caption{Generalized Gaussian density of accuracy of LFGCN (two blue dashed lines represent lower and upper bounds, respectively): (a) the red filled curve is the PR-based method ($\sigma = 0$), blue filled curve is the NL-based method ($\sigma = 0.5$), and green filled curve is the SL-based method ($\sigma = 1$). (b) the red, blue, green filled curves represent scenarios with fractional parameters $\gamma$ of 0.001, 0.005, and 0.010, respectively.\label{fig2}}
\end{center}
\end{figure}

\textbf{5.\enskip Hyperparameter sensitivity} In the sensitivity analysis setting, we have the ability to analyze the sensitivity of the node classification accuracy to variation from three LFGCN-specific parameters -- $\alpha \in \{0.1, \cdots, 1\}$, $\sigma \in \{0, 0.1,\cdots, 1\}$, and $\gamma \in \{0.001, 0.01, 0.1, 1\}$. In this case, we only show the results from sensitively analysis for LFGCN model on IEEE 118-bus dataset. First, we perform the parameter learning experiments on four scenarios with a fixed parameter $\gamma$. Figure~\ref{different_gamma_plots} shows that the accuracy substantially decreases when $\alpha$ is larger than 0.8, especially in $\gamma$ equals to 0.001 and 0.01 (see Figure~\ref{different_gamma_plots}(a),~\ref{different_gamma_plots}(b)). Setting $\gamma = \{0.1, 1\}$, we observe that the classification accuracy nearly monotonic decreases while increasing $\alpha$. Additionally, LFGCN generally gives consistent and higher accuracy for $\gamma = \{0.001, 0.01\}$ when the $\alpha$ parameter is within the range of \{0.1, 0.2, 0.3, 0.4\}. We then explore the variation of accuracy based on tuning parameter $\gamma$ within the range of $[0.001, 0.002, \cdots, 0.01]$ (setting $\sigma \in \{0, 0.1, \cdots, 1\}$ at the same time), however, it is hard to obtain the optimal $(\hat{\sigma}, \hat{\gamma})$ combination through gathering finite experimental results (100 runs) since some of the results are very close. Therefore, we run the following experiments to demonstrate the impact evaluation of $ \gamma$:
\begin{itemize}[leftmargin=*]
    \item \textit{Step 1:} Set $\gamma \in \{0.001, 0.002, \cdots, 0.1\}$.
    \item \textit{Step 2:} For each fixed $\gamma$, 
    we run our proposed model 100 times separately for $\sigma$ from 0 to 1 by 0.01. We obtain 100 average accuracies for each $\gamma$: $\{\text{Acc}_{\sigma = 0}, \text{Acc}_{\sigma = 0.01}, \cdots, \text{Acc}_{\sigma = 1}\}^{[i]}$, $1\leq i\leq 10$.
    \item \textit{Step 3:} Fit the Gaussian distribution to $\{\text{Acc}_{\sigma = 0}, \text{Acc}_{\sigma = 0.01}, \cdots, \text{Acc}_{\sigma = 1}\}^{[i]}$ (see Fig.~\ref{fig2}(b)). 
\end{itemize}

Similar to $\gamma$, we fit the Gaussian distribution to $\{\text{Acc}_{\gamma = 0.001}, \text{Acc}_{\gamma = 0.002}, \cdots, \text{Acc}_{\gamma = 0.01}\}^{[j]}$ by fixing $\sigma$, where $j = 1, \cdots, 11$ (see Figure~\ref{fig2}(a)).

Figure~\ref{fig2} shows that there exists a more profound difference between the shapes of approximate Gaussian distributions by fixing the parameter $\sigma$ than fixing the parameter $\gamma$. These findings imply that $\sigma$ tends to be a more important factor in the LFGCN approach for small datasets.


\section{Conclusion}

We have proposed a new L\'evy Flights Graph Convolutional Network (LFGCN) method for semi-supervised learning on graphs that enables to better capture the intrinsic local graph topology. 
In addition, to further mitigate over-fitting and over-smoothing, we have proposed a new preferential P-DropEdge algorithm, based on censoring higher edge betweenness order statistics. We have investigated theoretical properties of LFGCN  and have validated utility of individual components of the LFGCN architecture.

Our numerical studies have indicated that the new LFGCN method tends to outperform all competing deep learning approaches on both unweighted-directed and unweighted-undirected graphs in all considered datasets, except of PubMed.
The gain in learning accuracy of LFGCN over the next best competitor ranges from 0.29\% to 4.27\%, and the highest gain has been achieved for the IEEE 118-Bus dataset which is the smallest among the considered datasets. Furthermore, in contrast to the competing approaches, LFGCN tends to deliver a more stable performance across directed and undirected networks regardless of the label rate.

In the future we plan to advance the proposed LFGCN technique to learning on multilayer networks, explore utility of P-DropEdge combined with other order statistics on graphs, and  enhance graph learning process with topological information on the underlying deep neural network.

\bibliography{example_paper}
\bibliographystyle{IEEEtran}
\section*{Appendix}
\section{Proof of Lemma 1}
\label{lemma3_proof}

\textit{Proof.} The objective function of the generalized semi-supervised learning framework (i.e., Lemma~\ref{gssl_formula}) can be rewritten in the following matrix form:
 $Q(F)=2 \sum_{k=1}^{K} F_{. k}^{T} D^{\sigma-1} L D^{\sigma-1} F_{. k}+ \mu \sum_{k=1}^{K}\left(F_{. k}-Y_{. k}\right)^{T} D^{2 \sigma-1}\left(F_{. k}-Y_{. k}\right)$.
Given the first order optimality condition $D_{F. k} Q(F)=0$, we have
  $2 F_{. k}^{T}\left(D^{\sigma-1} L D^{\sigma-1}+D^{\sigma-1} L^{T} D^{\sigma-1}\right)
  +2 \mu\left(F_{. k}-Y_{. k}\right)^{T} D^{2 \sigma-1} =0$. 
Multiplying the above expression from the right hand side by $D^{-2 \sigma+1}$ leads to
  $2 F_{. k}^{T}\left(D^{\sigma-1}\left(L+L^{T}\right) D^{-\sigma}\right)+2 \mu\left(F_{. k}-Y_{. k}\right)^{T}=0$.
Then, substituting $L = D - W$ and rearranging the terms yields
    $F_{. k}^{T}\left(2 I-D^{\sigma-1}\left(W+W^{T}\right) D^{-\sigma}+\mu I\right)-\mu Y_{. k}^{T}=0$.
Since the resulting adjacency matrix $W$ is symmetric, 
    $F_{. k}^{T} =\mu Y_{. k}^{T}\left(2 I-2 D^{\sigma-1} W D^{-\sigma}+\mu I\right)^{-1}$.
\hfill $\Box$

\section{Proof of Lemma 2}
\label{lemma1_proof}
\textit{Proof.} Relaxation time of a Markov process is given by the reciprocal of the spectral gap $\min \{\lambda_2,2-\lambda_n \}$, where
$0=\lambda_1\le \lambda_2\le...\le \lambda_n$ are eigenvalues of the normalized Laplacian. From the singular value decomposition $L^{\gamma} = U^{\top}\Lambda^{\gamma} U$ and monotonicity of the power function, we conclude that $\lambda_2^\gamma$ and $\lambda_n^\gamma$ are the smallest and largest eigenvalues of the L\'evy flight, respectively. Since for $0<\gamma<1$, $x^\gamma > x$
when $0<x<1$ and $x^\gamma < x$
when $1<x<2$, the spectral gap of the L\'evy flight increases with respect to the original random walk. Consequently, the relaxation time of the L\'evy flight is less than the relaxation time of the original random walk. \hfill $\Box$

\section{Proof of Lemma 3}
\label{lemma2_proof}


\textit{Proof.} The parallel structure~\citep{henley1981reliability} is constructed in a similar spirit as bagging of features in random forest and other ensemble learning methods. The key rationale behind the parallel structure is to reduce variance and increase stability. Let us first consider the first equation in Lemma 2. Suppose that  $\Upsilon$ is a random sample drawn without replacement from a finite population $\mathcal{U}=\{1,2,\dots, N\}$ according to a sampling design $p(\cdot)$. Each unit $i, i=1,\ldots, N$ of $\mathcal{U}$ is associated with $X_{FGS}^{(i)}$, i.e. the $i$-th output matrix from FGS convolution for new feature matrix $X_i$. Probability of choosing sample $\upsilon$ is
$Pr(\Upsilon = \upsilon) = p(\Upsilon)>0$ for all $\upsilon \in U$ and $\upsilon \neq \emptyset$. Let $Z$ be the indicator variable such that
$ Z_{i}=1$ if $X_i$ is in the sample, and 0 otherwise.
Hence, probability that the unit $i,i=1,\ldots, N$ is selected, is given by $\pi_i=E(Z_{i})$,
and probability
that units $i$ and $j$, $i,j=1,\ldots, N$ are selected simultaneously is  $\pi_{ij}=E(Z_{i}Z_{j})$; 
$\pi_i$ and $\pi_{ij}$ are called {\it first order inclusion probability}
and {\it second order inclusion probability}, respectively.
Since in the current paper, we consider a simple random sampling design of $n$ units $X_{FGS}^{(i)}$ without replacement from $\mathcal{U}$, $\pi_i=n/N$ and $\pi_{ij}=n(n-1)/N(N-1)$. Let $\bar{X}_{FGS}=\sum_{i \in N}{X_{FGS}^{(i)}}/{n}=\sum_{i=1}^{N} Z_{i} {X_{FGS}^{(i)}}/{n}$. Hence, given $\{X_{FGS}^{(1)}, \ldots, X_{FGS}^{(N)}\}$, we find that
\begin{eqnarray*}
    E[\bar{X}_{FGS}]=\sum_{i=1}^{N} E\left[Z_{i}\right] \frac{X_{FGS}^{(i)}}{n}
    = \sum_{i=1}^{N} \pi_i \frac{X_{FGS}^{(i)}}{n}
     =\bar{X}_{FGS}^{U},
\end{eqnarray*}
where $\bar{X}_{FGS}$ is be the output matrix $P_1$ from a pooling layer (see Residual Blocks in main text Figure 1), which implies that
$\bar{X}_{FGS}$ is an unbiased estimator of population mean $\bar{X}_{FGS}^U$.


In turn, 
\begin{equation*}
\begin{split}
   &\operatorname{Var}(\bar{X}_{FGS}) 
    = \frac{1}{n^{2}}\left[\sum_{i=1}^{N} (X_{FGS}^{(i)})^2 \operatorname{Var}\left(Z_{i}\right) \right.\\
    &\left. +\sum_{i=1}^{N} \sum_{j \neq i}^{N} X_{FGS}^{(i)} X_{FGS}^{(j)} \operatorname{Cov}\left(Z_{i}, Z_{j}\right)\right]
      =\left(1-\frac{n}{N}\right) \frac{S^{2}}{n},
\end{split}
\end{equation*}
where $S^{2}=\sum_{i=1}^{N}\left(X_{FGS}^{(i)}-\bar{X}_{FGS}^{U}\right)^{2} /(N-1)$. Factor $(1 - n/N)$ is called the finite population correction (FPC). Hence, the proposed PS structure decreases an original estimation variance $S$ by order of $n$.
Note that we can consider other sampling designs $p(\cdot)$, including, for example, weighted sampling and $p$-extended simple random sampling with replacement~\citep[see discussion by][and references therein]{wong1980efficient, scott1994sampling, ozturk2019constructing}. \hfill $\Box$ 

The stability result of the parallel structure~\citep{nowak2012reliability} follows from verbatim application of the probability bound on the intersection of independent events.   

\section{More Details on Experiments}
\label{experiments_details}
The statistics of data we used in the experimental section are summarized in Table~\ref{data_stat}. For undirected networks and Cora, we trained and tested our model on the same dataset splits as in~\citep{kipf2017}; and we use a $10\%$/$20\%$/$70\%$ split into training, validation, and test sets for power grid networks.
\begin{table}[h]
\caption{Dataset statistics.}
\label{data_stat}
\begin{center}
\begin{tabular}{lccccc}
\toprule
\multicolumn{1}{l}{\bf Dataset}  &{\bf Vertices}&{\bf Edges}&{\bf Features}&{\bf Classes}&{\bf Label rate}
\\ \hline 
Cora-ML&2,708&5,429&1,433&7&0.052\\
CiteSeer&3,327& 4,732&3,703&6&0.036\\
PubMed&19,717&44,338&500&3&0.003\\
Cora& 19,793& 65,311 & 8,710 & 70&0.100 \\
IEEE Bus&118&182&2&3&0.100\\
TX Bus&2,000&2,668&5&3&0.100\\
SC Bus&500&584&5&3&0.100\\
\bottomrule
\end{tabular}
\end{center}
\end{table}

We report results for the mean training time per epoch of LFGCN for both undirected and directed networks on Tesla V100-SXM2-16GB.

\begin{table}[H]
\caption{Time per epoch for LFGCN training on both undirected and directed networks.}
\label{sample-table}
\begin{center}
\scalebox{0.85}{
\begin{tabular}{lccccccc}
\toprule
{\bf Dataset} & {\bf Cora-ML} & {\bf CiteSeer} & {\bf PubMed} & {\bf Cora} & {\bf IEEE} & {\bf TX} & {\bf SC}\\
\midrule
Runtime& 0.04s  &0.27s&  0.22s& 0.15s& 0.01s&  0.14s&   0.02s\\
\bottomrule
\end{tabular}
}
\end{center}
\vskip -0.1in
\end{table}

All high-order approaches, either fractional or integer powers of Laplacian, lead to increased computational costs. Table~\ref{run_time2} shows running times for 5 SOTAs on Citeseer. LFGCN time is compatible with SOTAs. 
\begin{table}[h]
\caption{Runtime on CiteSeer.}
\label{run_time2}
\begin{center}
\begin{tabular}{lccccc}
\toprule
 {\bf Method}&{\bf GCN}&{\bf MixHop}&{\bf VPN}&{\bf APPNP}&{\bf LFGCN}
\\ \hline 
Runtime &0.22s& 0.41s&0.26s&0.27s&0.27s\\
\bottomrule
\end{tabular}
\end{center}
\vskip -0.1in
\end{table}

To highlight comparison of DropEdge vs. P-DropEdge, we show runtime/per epoch of LFGCN + DropEdge vs. LFGCN + P-DropEdge for Citeseer (edge betweenness is computed offline): 

\begin{table}[h]
\caption{Runtime of LFGCN with two dropedge methods.}
\label{run_time3}
\begin{center}
\begin{tabular}{lcc}
\toprule
 {\bf Method}&{\bf LFGCN + DropEdge}&{\bf LFGCN + P-DropEdge}
\\ \hline 
Runtime &0.28s& 0.33s\\
\bottomrule
\end{tabular}
\end{center}
\vskip -0.1in
\end{table}

Note that while calculating edge betweenness may be time consuming, it's performed offline. Different from DropEdge, P-DropEdge is an offline method based on prior edge info and such prior info is not limited to edge betweenness. Instead, we can use, e.g. edge criticality based on percolation theory, k-path edge centrality with time complexity $O(k|E|)$, and centrality based on the edge impact on giant comp -- scores particularly important in weighted graphs, e.g power and transportation systems. P-DropEdge can bring new insights on how targeted perturbation of graph topology can assist addressing GCN oversmoothing and GCN sensitivity to attacks.


\end{document}